\documentclass[10pt,twocolumn,letterpaper]{article}

\usepackage{enumitem}
\usepackage[pagenumbers]{cvpr} 

\usepackage[dvipsnames]{xcolor}

\definecolor{cvprblue}{rgb}{0.21,0.49,0.74}
\usepackage[pagebackref,breaklinks,colorlinks,citecolor=cvprblue]{hyperref}

\usepackage{booktabs}
\usepackage{multirow}

\usepackage{subcaption}

\usepackage{flushend}

\usepackage[sectionbib]{chapterbib}

\usepackage[normalem]{ulem}
\useunder{\uline}{\ul}{}

\def\paperID{**} 
\def\confName{**}
\def\confYear{2024}

\title{Multiple View Geometry Transformers for 3D Human Pose Estimation}

\author{Ziwei Liao\textsuperscript{1,*} \quad Jialiang Zhu\textsuperscript{2,*}  \quad Chunyu Wang\textsuperscript{3,†}  \quad Han Hu\textsuperscript{3}  \quad Steven L. Waslander\textsuperscript{1}\\
\textsuperscript{1} University of Toronto \quad \textsuperscript{2} Southeast University \quad \textsuperscript{3} Microsoft Research Asia\\
{\tt\small ziwei.liao@mail.utoronto.ca, steven.waslander@robotics.utias.utoronto.ca} \\ 
{\tt\small jialiangzhu@seu.edu.cn  }\\
{\tt\small chnuwa@microsoft.com, hanhu@microsoft.com }
}

\begin{document}
\maketitle

{\let\thefootnote\relax\footnote{{* Equal contribution}}}
{\let\thefootnote\relax\footnote{{† Correspondance author}}}

\begin{abstract}

In this work, we aim to improve the 3D reasoning ability of Transformers in multi-view 3D human pose estimation. Recent works have focused on end-to-end learning-based transformer designs, which struggle to resolve geometric information accurately, particularly during occlusion. Instead, we propose a novel hybrid model, \emph{MVGFormer}, which has a series of geometric and appearance modules organized in an iterative manner. The geometry modules are learning-free and handle all viewpoint-dependent 3D tasks geometrically which notably improves the model's generalization ability. The appearance modules are learnable and are dedicated to estimating 2D poses from image signals end-to-end which enables them to achieve accurate estimates even when occlusion occurs, leading to a model that is both accurate and generalizable to new cameras and geometries. We evaluate our approach for both in-domain and out-of-domain settings, where our model consistently outperforms state-of-the-art methods, and especially does so by a significant margin in the out-of-domain setting. We will release the code and models: \url{https://github.com/XunshanMan/MVGFormer}.
\end{abstract}    
\section{Introduction}

\begin{figure}
    \centering
    \includegraphics[width=0.45\textwidth]{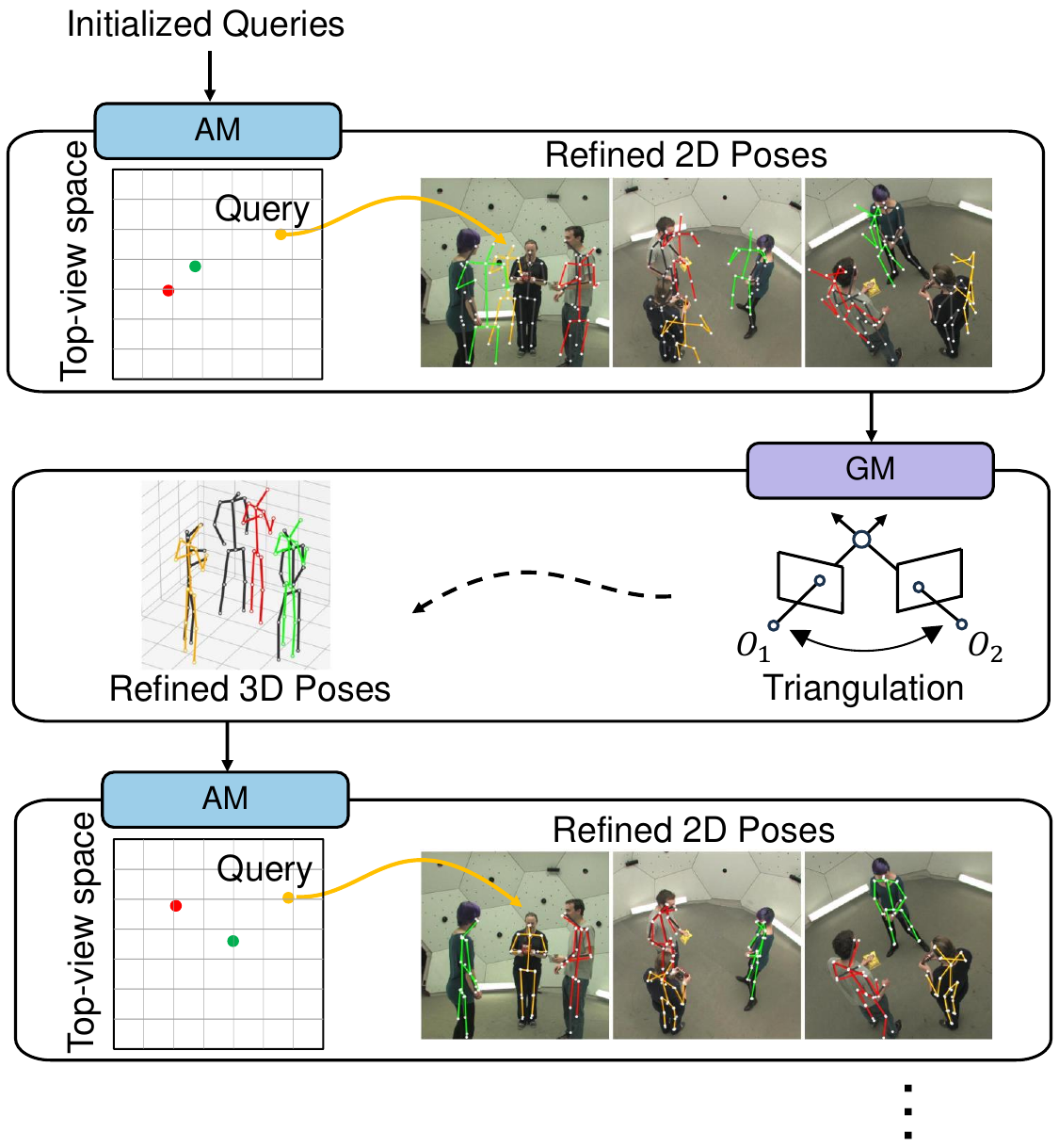}
    \vspace{-0.3cm}

    \caption{Overview of our MVGFormer which directly introduces multiple view geometry reasoning in Transformers. It is fully-differentiable and end-to-end trainable. The two types of modules (AM \& GM) iteratively refine the 2D and 3D poses, respectively.}
    \label{fig:framework_new}

    \vspace{-0.5cm}
\end{figure}

Estimating 3D human pose from multiple cameras is an important task with diverse applications.
Recently, state-of-the-art accuracy on benchmark datasets~\cite{ionescu2014human3,Joo_2017_TPAMI} has been advanced significantly by the application of large neural networks~\cite{tu2020voxelpose,zhang2021direct,reddy2021tessetrack}, due to their capabilities to learn complex priors required to handle severe occlusion. However, our study shows that the learnable prior is a double-edged sword that may harm its generalization ability, especially in the multiple-view setting where the camera views may undergo large viewpoint variations in an environment unseen during training. This observation inspires us to consider the following question in our work:

\vspace{0.5em}

\emph{What sub-tasks should be addressed geometrically in a learning-free style rather than by neural networks?}

\vspace{0.5em}

The core challenge of multiple view 3D human pose estimation is to establish correspondence across views and reason about the 3D world from 2D measurements. Some methods~\cite{dong2019fast, dong2021fast, zhang20204d, perez2022matching, bridgeman2019multi} address the task from a primarly geometric direction. They first estimate 2D poses in each view, group them into clusters, and then estimate a 3D pose for each cluster by triangulation. They generalize well to unseen settings because of explicit geometry modeling. However, they struggle to handle severe occlusion due to the lack of effective priors leading to correspondence failures, when compared to learning-based neural networks. 

\vspace{0.3em}
Recent works~\cite{zhang2021direct,iskakov2019learnable,tu2020voxelpose,reddy2021tessetrack,ye2022faster} have emerged that rely on deep networks such as Transformers to map implicit image features to 3D poses bypassing explicit geometry modeling. Some methods such as~\cite{zhang2021direct} even map camera parameters to latent features for end-to-end training. The learned priors embedded in the networks enable them to achieve accurate estimates even when occlusion occurs.  However, as will be shown in our experiments, learned approaches tend to obtain poor results on new scenes, especially when the testing camera configurations are not seen in training.

\begin{figure*}
    \centering
    \includegraphics[width=0.98\textwidth]{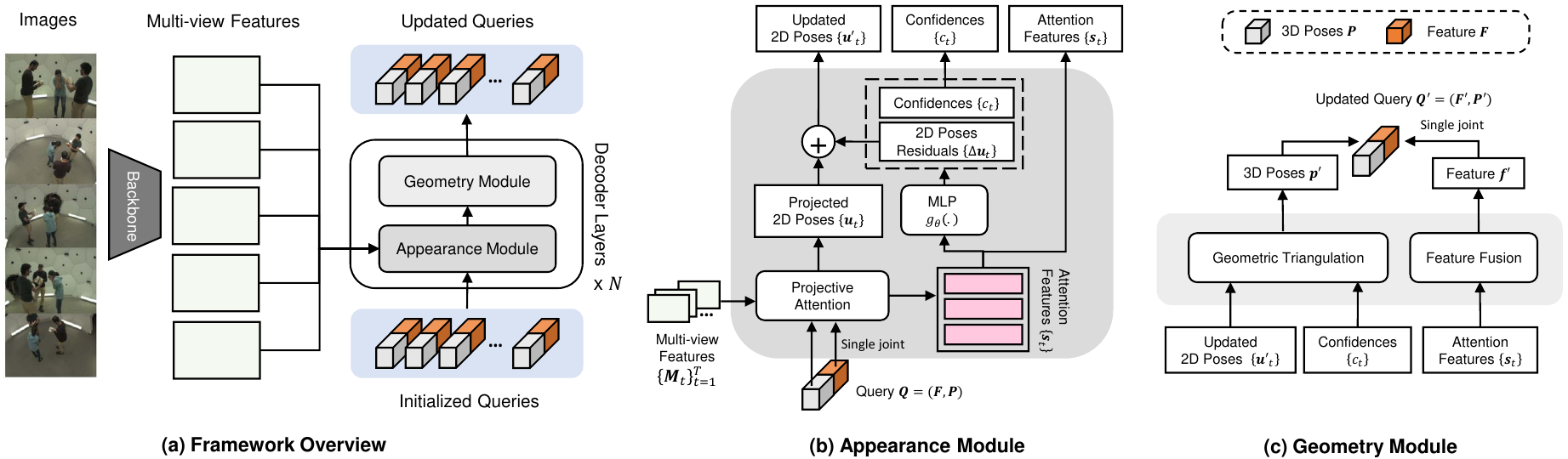}
    \setlength{\belowcaptionskip}{-12pt}
    \caption{(a) Overview of MVGFormer which estimates 3D human poses from multiple cameras with \textit{N} decoder layers. (b) The appearance module updates the 2D poses in each view mainly dependent on image features. (c) The geometry module explicitly estimates the 3D poses based on the previously estimated 2D poses and the camera parameters. The two modules iteratively refine the initialized queries in a coarse-to-fine manner.}
    \label{fig:framework}
\end{figure*}

In this paper, we present a novel hybrid model called MVGFormer that combines Multiple View Geometry reasoning with Transformers. As shown in Fig. \ref{fig:framework_new} and Fig. \ref{fig:framework} (a), it takes as input a set of initialized queries that encode 3D poses, and iteratively refines them using two modules, the Appearance Module (AM) and Geometry Module (GM). In AM, we project a query onto the images to get a set of sampling points, and estimate a more accurate 2D pose from the sampled features (see Fig. \ref{fig:framework} (b)). With the refined 2D poses in all views, GM predicts a refined 3D pose in a learning-free style through triangulation, which in turn is used to update the query (see Fig. \ref{fig:framework} (c)). Thus, in the next round, AM can sample features at more accurate positions, leading to a refined 2D pose.

\vspace{0.5em}
\emph{MVGFormer} has four advantages. First, compared with a pure learning-based method that directly regresses 3D poses and lacks the geometric understanding of the 2D-3D relationship, we explicitly introduce the 3D geometry module, simplifying and constraining the learning task. This enhances the coordination between the appearance and geometry modules across multiple decoder layers. Second, decomposing the 2D and 3D tasks reduces the complexity of the problem. Third, a query, which represents a single 3D pose, geometrically guides the system to relate, fuse, and improve the 2D estimates in each view. Refining queries in a coarse-to-fine manner implicitly solves the challenging cross-view person grouping stage, by addressing the problem from a top-down perspective, making it robust to low-level errors from occlusion. Furthermore, during the end-to-end training, the geometric module does not suffer from quantization errors faced by voxel-based methods~\cite{tu2020voxelpose,zhang2022voxeltrack}.

\vspace{0.5em}
We evaluate \emph{MVGFormer} on three benchmark datasets in two settings: the commonly used in-domain setting and the less studied out-of-domain setting, where the method is evaluated on a dataset with viewpoints not seen during training. The latter is used to show the generalization ability of different models. We observe that our model is able to achieve accurate 2D estimates even when severe occlusion occurs (see Fig. \ref{fig:exp:cmu} for examples) due to the effective multi-view regularization in GM during iterative refinement. This lays the foundation for triangulation to obtain accurate 3D poses. Our method outperforms the state-of-the-art methods consistently on the three datasets. Further, it gets notably better results in the out-of-domain setting.

\section{Related Work}
\noindent\textbf{Geometry modeling}. Triangulation 
is commonly used in multi-view pose estimation~\cite{iskakov2019learnable, dong2019fast, dong2021fast, chen2022structural, bartol2022generalizable, wu2021graph, perez2022matching, zhang20204d}. It takes as input the pre-detected 2D poses and computes a 3D pose by multi-view geometry~\cite{hartley2003multiple}. 
Some work assumes only a single person in the scene~\cite{iskakov2019learnable, chen2022structural, bartol2022generalizable}. When multiple persons exist, data associations across views need to be resolved. 
One group of works~\cite{dong2019fast,dong2021fast,perez2022matching} follows a two-stage design, first solving associations with a hand-crafted affinity matrix, and then estimating 3D poses with triangulation. Though generalized to any cameras, they are vulnerable to inaccurate 2D pose estimates. 
A further set of methods require learning-based modules for data associations~\cite{zhang20204d,wu2021graph}, pose refinements~\cite{wu2021graph}, or depth regression~\cite{lin2021multi}, which face generalization challenges since the learning process depends on the training cameras. Most recently, Bartol et al.~\cite{bartol2022generalizable} investigate the generalization performance of triangulation, but only consider a single person in the scene. 
Video input has also been exploited for 3d pose estimation~\cite{zhang2022voxeltrack,reddy2021tessetrack,zhang20204d,dong2021fast} relying on temporal information to improve pose tracking over time.  
In this work, we focus on single-frame results, leaving the tracking problem as future work.

\noindent\textbf{Volumetric Representation} is another way to introduce geometric information into neural networks by projecting 2D image features into 3D~\cite{pavlakos2017coarse,pavlakos2018ordinal}. VoxelPose~\cite{tu2020voxelpose} and its variants~\cite{ye2022faster,reddy2021tessetrack,zhang2022voxeltrack,chen2022vtp} are more robust to occlusion by fusing multi-view image features in a voxelized 3D representation. However, the high computational complexity and memory requirements of the dense 3D convolutions lead to a discretization tradeoff that frequently results in large quantization errors of ~$30$mm~\cite{tu2020voxelpose} to fit on available hardware. VoxelPose is also susceptible to increased errors when applied to novel cameras or datasets, due to the learned dependence of visual patterns that arise in the voxels for specific camera angles observed during training. Performing extensive augmentations such as voxel rotations~\cite{iskakov2019learnable} is needed to help alleviate the problem, but the challenge persists.

\noindent\textbf{Transformers for 3D Detections}. DETR~\cite{carion2020end} is a Transformer framework for 2D image tasks and has been extended to 3D object detection from point clouds~\cite{misra2021end,he2022voxel,fan2022embracing}. Some works~\cite{li2022bevformer,yang2022bevformer,liu2022bevfusion,zhang2022beverse} construct volumetric BEV representations from multi-view images for 3D object detection in autonomous driving scenarios. However, they are not suitable for 3D human pose estimation because of the large quantization errors. PETR~\cite{liu2022petr, liu2022petrv2} encodes the position information of 3D coordinates as features into DETR for multi-view 3D object detection. However, it ignores the geometry constraints and requires a large amount of data to train the network to accurately learn the 2D-to-3D relationship. Thus, the approach results in overfitting to the training camera configurations, which limits its generalization performance. 
Most closely related to our work, MvP~\cite{zhang2021direct} extends DETR for multi-view 3D human pose estimation. Similar to PETR~\cite{liu2022petr, liu2022petrv2}, it introduces a RayConv operation to integrate the camera parameters into the image features. Although it achieves good in-domain performance when cameras are the same for training and testing, it does not generalize well to different camera arrangements. In our experiments, the accuracy almost decreases to zero when we apply the model to different camera arrangements or datasets. Taking our insight from these stark results on out-of-domain performance for existing learned methods such as MvP, we present a principled approach to leverage DETR in multi-view 3D tasks by enabling explicit multi-view geometry modeling within a course-to-fine refinement pipeline.

\section{Methodology} 

A schematic illustration is provided in Figure~\ref{fig:framework}. The input is a set of images ${\{\mathbf{I}_t\}}_{t=1}^{T}$ collected from $T$ cameras.  The images are passed to a backbone network to extract the multi-view feature maps ${\{\mathbf{M}_t\}}_{t=1}^{T}$.  We initialize \textit{K} compositional queries ${\{\mathbf{Q}_k\}}_{k=1}^{K}$ (Section \ref{section:query}), which go through a series of decoder layers to be iteratively refined to estimate the 3D poses and features (Section \ref{section:decoder}) for each person in the scene. We present the training objectives in Section \ref{sec:training}.

\subsection{Compositional Query}
\label{section:query}
We use a compositional query to represent an individual person in the scene. Each query $\mathbf{Q}_k$ consists of an appearance term $\mathbf{F}_k \in \mathbb{R}^{J \times L}$ and a geometry term $\mathbf{P}_k \in \mathbb{R}^{J \times 3}$ where $J$ and $L$ denote the number of joints and the feature dimension, respectively. Note that each body joint has a separate feature vector in both terms. The two terms form the compositional query: $\mathbf{Q}_k=\left(\mathbf{F}_k, \mathbf{P}_k \right)$.

\vspace{0.5em}
\noindent
\textbf{Appearance Term } We adopt a hierarchical query embedding scheme proposed in~\cite{zhang2021direct} to reduce the number of learnable parameters. In particular, we learn $K$ instance embeddings $\{\mathbf{h}_k\}_{k=1}^{K} \subset \mathbb{R}^L$ and $J$ joint embeddings  $\{\mathbf{g}_j\}_{j=1}^{J} \subset \mathbb{R}^L$. The appearance term of joint $j$ of the $k^{\text{th}}$ query is obtained by $\mathbf{f}_k^j=\mathbf{h}_k+\mathbf{g}_j$. The appearance terms of all joints are denoted as: $\mathbf{F}_k=[\mathbf{f}_k^1,\mathbf{f}_k^2,...,\mathbf{f}_k^J]^\intercal$. The embeddings are randomly initialized from the standard normal distribution during training.

\begin{figure*}
    \centering
    \includegraphics[width=0.95\textwidth]{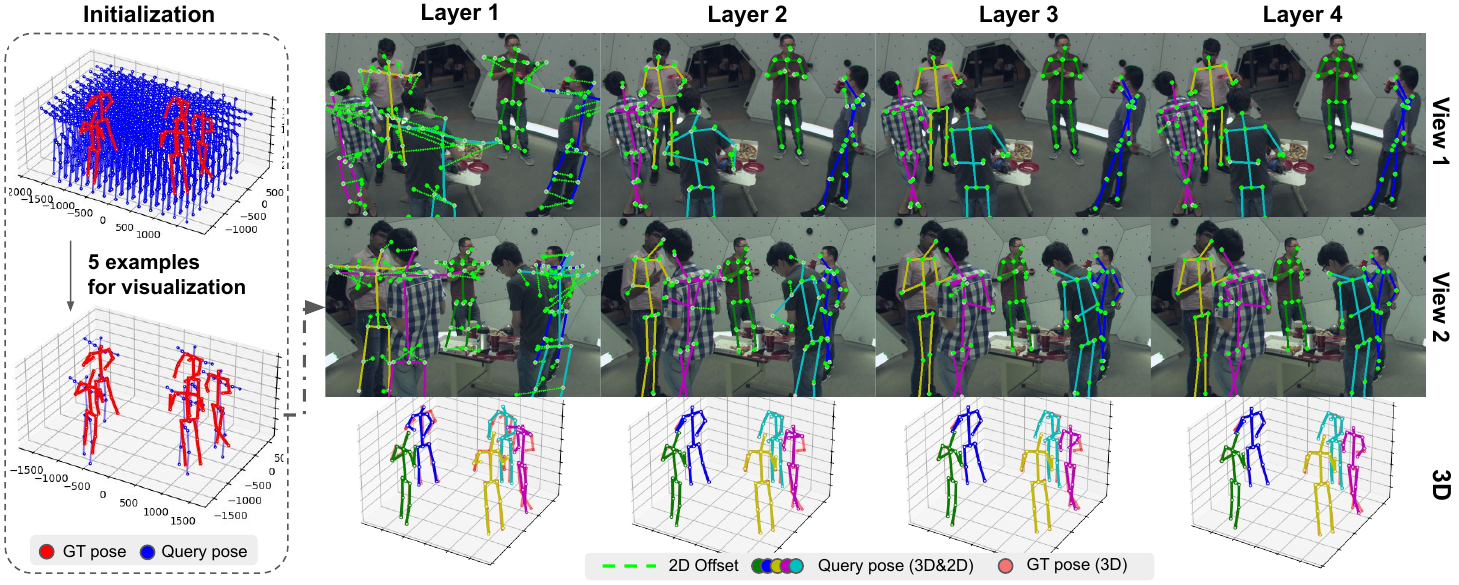}
    \caption{Iterative query refinement through decoder layers. (1) Initialization. We uniformly sample $N$ T-poses on the ground to initialize queries. We visualize the details of 5 queries after filtering for clarity. (2) For each query, each joint is projected into 2D images to aggregate features with attention, and is updated with a learnt offset (green dashed line). Then, triangulation is used to estimate a 3D pose based on updated 2D poses in each view. The process is repeated in the four layers. 
    Note that GT poses are NOT used during inference and are for visualization only. We recommend viewing electrically in color.}
    \label{fig:real_example_framework}

    \vspace{-0.5cm}
\end{figure*}

\vspace{0.5em}
\noindent
\textbf{Geometry Term} The geometry term $\mathbf{p}_k^j \in \mathbb{R}^3$ of joint $j$ of the $k^{\text{th}}$ query directly stores its 3D position. We leverage a simple initialization method to ensure the best generalization. We place a T-pose at $K$ uniformly sampled human centers on the ground plane and use the resulting 3D positions of the body joint to initialize $\mathbf{p}_k^j$, as in Figure~\ref{fig:real_example_framework} and Figure ~\ref{fig:distributions_queries}. The geometry terms can also be more accurately initialized by the 3D pose predictions obtained by other methods, like VoxelPose of which we will show examples in experiments. The geometry terms of all joints of the $k^{\text{th}}$ query are denoted as: $\mathbf{P}_k=[\mathbf{p}_k^1,\mathbf{p}_k^2,...,\mathbf{p}_k^J]^\intercal$.

\subsection{Transformer Decoders}

\label{section:decoder}
As shown in Figure \ref{fig:framework} (a), the model has multiple transformer decoder layers, with each having two successive modules that focus on the 2D and 3D tasks, respectively. We formally introduce the two modules in the following subsections.
For the sake of clarity, we slightly abuse the notation and use $\mathbf{f}$ and $\mathbf{p}$ to represent the appearance and geometry terms of a single joint, respectively, and omit the subscripts $k$ and $j$.

\paragraph{Appearance Module (AM)} For each query, AM is responsible for refining the corresponding 2D poses in all views, respectively. Figure~\ref{fig:framework} (b) shows the architecture.  For each query, the projective attention module extracts the attention features in each view for refining its 2D position. In particular, we project $\mathbf{p}$ to view \textit{t} and obtain the 2D position $\mathbf{u}_t$: $\mathbf{u}_t=\mathbf{\Pi}_t[\mathbf{p} ; 1]$, where $\mathbf{\Pi}_t \in \mathbb{R}^{3 \times 4}$ is the known projection matrix of camera ${t}$ and $[\mathbf{p} ; 1]$ is the homogeneous form.
Then, following the deformable attention~\cite{zhu2020deformable}, we extract an attention feature vector $\mathbf{s}_t$ from the feature map ${\mathbf{M}_t}$ by predicting a set of sampling locations around $\mathbf{u}_t$. Then we learn an MLP $g_\theta(.)$ to predict a 2D residual (offset) $\Delta\mathbf{u}_t$ and a confidence score $c_t$ from $\mathbf{s}_t$:
\begin{equation}
    [\Delta\mathbf{u}_t,c_t] = g_\theta(\mathbf{s}_t),
\label{eq:2doffset}
\end{equation}
where $c_t$ denotes the estimation confidence, used by the geometry module to reduce the impact of inaccurate 2D poses on the triangulation. We supervise $c_t$ indirectly by minimizing the errors of the estimated 3D poses since the whole process is differentiable. We supervise $\Delta\mathbf{u}_t$ using the projected groundtruth 2D poses. We get the refined 2D position by adding the projected position and the residual: 
\begin{equation}
 \mathbf{u'}_t=\mathbf{u}_t + \Delta\mathbf{u} _t  
\end{equation}
Besides estimating the residuals, we also use the attention feature $\mathbf{s}_t$ to update the query feature vector $\mathbf{f}$ as will be stated in the subsequent section. The updated query feature can better reflect the characteristics of the corresponding people in the images.

\paragraph{Geometry Module (GM)}
This module aims to estimate a more accurate 3D pose from the refined 2D poses. Figure~\ref{fig:framework} (c) shows the architecture. With the 2D positions in $T$ views $\{\mathbf{u'}_t\}_{t=1}^T$, and the scores $\{c_t\}_{t=1}^T$ predicted in the previous module, we compute a new 3D position $\mathbf{p}'$ by differentiable algebraic triangulation~\cite{iskakov2019learnable}: 
\begin{equation}
    \mathbf{p}' = \text{Triangulate}(\{\mathbf{u'}_t\}_{t=1}^T, \{\mathbf{c}_t\}_{t=1}^T, \{\mathbf{\Pi_t}\}_{t=1}^T),
\label{eq:triangulation}
\end{equation}
The scores are used to reduce the impact of inaccurate 2D poses. This module can be regarded as a denoising function because triangulation, which ensembles the multi-view 2D poses, is able to obtain a relatively accurate 3D pose estimate even when the 2D poses in a small number of views are inaccurate. Thus, projecting the 3D pose back to each view usually gives a more accurate 2D pose, from which the appearance module can attract more accurate and relevant features.

\begin{figure*}
    \centering
    \includegraphics[width=1.0\textwidth]{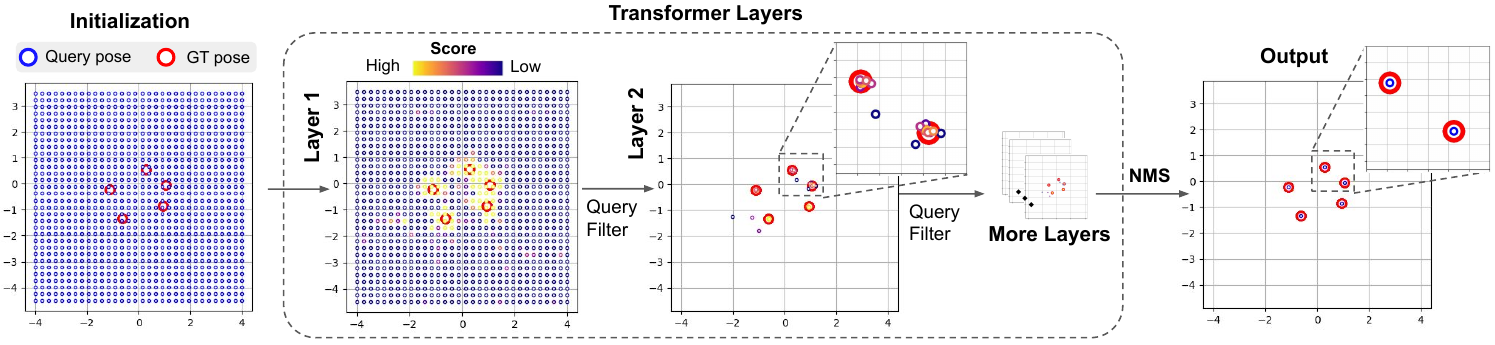}
    \vspace*{-5mm}
    \setlength{\belowcaptionskip}{-12pt}
    \caption{An example of the query distributions on the CMU Panoptic dataset with 5 persons. We show the center points of each pose from the top view. We uniformly sample $N=1024=64\times64$ points on the ground and then initialize a T-pose for each. For each decoder layer, we filter queries with scores below a threshold. Finally, we use NMS to keep only one query with the highest score from a cluster of queries attached to the same person. GT poses are not used during inference and are for visualization only. }
    \label{fig:distributions_queries}
\end{figure*}

Since the number of queries is larger than the actual number of people, we train an MLP-based classifier $f_{\beta}(.)$ to predict a score for each query based on the appearance term to remove the ``empty'' ones. It is set to be one for the positive queries and zero for the negative ones during training. We compute an overall score for each pose by averaging the scores of all joints. In both training and testing, we filter out the queries whose scores are smaller than $\epsilon$ layer by layer, as shown in Figure~\ref{fig:distributions_queries}. This helps spend computation only on the relevant regions in the space.

\vspace{0.5em}
Finally, we update each query for the next decoder layer. 
The geometry term is directly updated by the estimated 3D joint position $\mathbf{p}'$. The appearance term is updated by fusing the attention features in all views $\{\mathbf{s_t}\}_{t=1}^T$, which contains more accurate information describing the target person. We first compute a mean vector from $\{\mathbf{s_t}\}_{t=1}^T$. Then, we use an MLP $f_{\alpha}(.)$ to estimate a feature residual and add it to the origin appearance term $\mathbf{f}$. Finally, we feed it to another MLP $f_\gamma(.)$ to obtain $\mathbf{f}'$:
\begin{equation}
    \mathbf{f}' = f_\gamma(\mathbf{f} + f_{\alpha}(\text{Mean}(\{\mathbf{s_t}\}_{t=1}^T)))
\label{eq:featurefusion}
\end{equation}
It is worth noting that fusing multi-view features makes the approach more robust to occlusion in individual views, which improves the performance of our method. 

\subsection{Training Objectives}
\label{sec:training}
The whole model is end-to-end differentiable. We use an anchor-based matching strategy~\cite{zhou2019objects} to assign the queries to the ground truth poses. In particular, the initialized geometry terms of all queries $\{\mathbf{P}_k^{(0)}\}_{k=1}^K$ are regarded as the anchors. The ground-truth poses with $Z$ persons are denoted as $\{\mathbf{H}_z\}_{z=1}^Z \subset \mathbb{R}^{J \times 3}$. For each $\mathbf{H}_z$, we assign it to $W$ nearest queries $\{\mathbf{P}_{z(w)}\}_{w=1}^W$ based on their geometry terms where $z(w)$ is the index of the $w^{th}$ matched query. The unassigned queries are the negative anchors. The assignment is fixed during training. Our learning objective has two terms. The first is a pose loss supervising the 2D and 3D poses estimated from the positive queries:
\begin{multline}
L_{pose}(\{\mathbf{P}_k\},\mathbf{H}_z) = \sum _{w=1}^W ( 
\underbrace{L_1(\mathbf{P}_{z(w)}, \mathbf{H}_z)}_\textrm{3D pose loss} \\ + 
\sum_{t=1}^T \underbrace{L_1(\mathbf{\hat{U}}_{{z(w)},t}, \mathbf{U}_{z,t})}_\textrm{2D pose loss} )
\end{multline}
where $\mathbf{\hat{U}}_{{z(w)},t}$ represents the predicted 2D pose of the $z(w)^{\text{th}}$ query in the $t^{\text{th}}$ view and $\mathbf{U}_{z,t}$ represents the corresponding groundtruth 2D pose. The second is a cross-entropy loss for the classifier $f_{\beta}$ applied to both positive and negative queries. We enforce the above losses at every decoder layer.

\section{Experiments}
\subsection{Benchmarks and Implementation Details}
\textbf{Datasets, baselines, and metrics.} 
We evaluate our method in two settings: in-domain and out-of-domain on three common benchmarks, CMU Panoptic~\cite{Joo_2017_TPAMI}, Shelf~\cite{belagiannis20143d}, and Campus~\cite{belagiannis20143d}. We compare our method to three state-of-the-art baselines: two end-to-end learning methods: a Transformer-based MvP~\cite{zhang2021direct}, and a volumetric-based VoxelPose~\cite{tu2020voxelpose}; and a geometric method using triangulation Dong et al.~\cite{dong2019fast}.
We use their public official codes. Besides, we also list the published results of other works in the in-domain setting. 
Following~\cite{tu2020voxelpose,zhang2021direct}, we use metrics of Average Precision (AP), Recall, and Mean Per Joint Position Error (MPJPE) for the CMU Panoptic dataset, and Percentage of Correct Parts (PCP) for the Campus and Shelf datasets. Following the common practice, we use the camera poses provided by the datasets for a fair comparison of all the models. We discuss practical methods to estimate the camera poses in Appendix~\ref{sec:suppl:camera_parameters}.

\vspace{0.5em}
\noindent
\textbf{Implementation details}. Following previous work~\cite{tu2020voxelpose,zhang2021direct}, we use ResNet-50 as the backbone for extracting image features. It is pre-trained for 2D pose estimation~\cite{xiao2018simple} on the COCO dataset. Thereafter, the backbone is fixed in the subsequent training steps. The rest modules are jointly trained for 40 epochs with a learning rate of $4e^{-4}$. We use four decoder layers with independent parameters for each. 
We filter out the queries whose scores are smaller than $\epsilon=0.1$ at each layer. Finally, we use NMS to remove the redundant queries to obtain the final set of estimates, as shown in Figure~\ref{fig:distributions_queries}.
For the sampling space of the $1024$ initialized queries, we use the given studio space for the CMU Panoptic dataset and use the maximum range of all the persons in the training set for the Campus and Shelf dataset.
We present more architectural details in Appendix~\ref{sec:suppl:network_architectures}.

\subsection{Generalization Performance}
\label{section:generalization}

\begin{table*}[]
\centering

\resizebox{1.6\columnwidth}{!}{%

\begin{tabular}{c|cccccccccc}
\hline
\textbf{Generalization}     & \multicolumn{10}{c}{Change camera number}                                                                                                                                                                                                         \\ \hline
\textbf{Inference Datasets} & \multicolumn{2}{c|}{CMU0 (3)}                      & \multicolumn{2}{c|}{CMU0 (4)}                      & \multicolumn{2}{c|}{CMU0 (6)}                      & \multicolumn{2}{c|}{CMU0 (7)}                      & \multicolumn{2}{c}{Average}   \\ \hline
\textbf{Metrics}            & AP25          & \multicolumn{1}{c|}{mAP}           & AP25          & \multicolumn{1}{c|}{mAP}           & AP25          & \multicolumn{1}{c|}{mAP}           & AP25          & \multicolumn{1}{c|}{mAP}           & AP25          & mAP           \\ \hline
Dong et al.~\cite{dong2019fast}                        & 0.0           & \multicolumn{1}{c|}{37.7}          & 0.0           & \multicolumn{1}{c|}{46.6}          & 0.1           & \multicolumn{1}{c|}{41.8}          & 0.2           & \multicolumn{1}{c|}{40.1}          & 13.3          & 41.6          \\
VoxelPose~\cite{tu2020voxelpose}                   & 13.7          & \multicolumn{1}{c|}{73.0}          & 65.4          & \multicolumn{1}{c|}{93.1}          & 84.5          & \multicolumn{1}{c|}{96.8}          & 81.5          & \multicolumn{1}{c|}{96.3}          & 69.1          & 89.8          \\
MvP~\cite{zhang2021direct}                         & 12.3          & \multicolumn{1}{c|}{57.1}          & 59.8          & \multicolumn{1}{c|}{84.1}          & 0.0           & \multicolumn{1}{c|}{0.3}           & 0.0           & \multicolumn{1}{c|}{0.2}           & 21.6          & 35.4          \\
Ours                        & \textbf{44.6} & \multicolumn{1}{c|}{\textbf{83.4}} & \textbf{83.6}          & \multicolumn{1}{c|}{\textbf{96.0}}          & \textbf{94.7} & \multicolumn{1}{c|}{\textbf{98.5}}          & \textbf{95.1} & \multicolumn{1}{c|}{\textbf{98.6}} & \textbf{83.3} & \textbf{94.1} \\ \hline
\end{tabular}

}

\caption{Generalization experiments on \textbf{changing camera number}. All models are trained on CMU0 with 5 cameras, and tested on \textit{Inference Datasets}. CMU0 ($K$) means there are a total of $K$ cameras after removing or adding cameras.}
\label{table-generalization-camera_number}
\end{table*}

\begin{table*}[]
\centering

\resizebox{1.6\columnwidth}{!}{%
\begin{tabular}{c|cccccccccc}
\hline
\textbf{Generalization}     & \multicolumn{10}{c}{Change camera arrangements}                                                                                                                                                                                                   \\ \hline
\textbf{Inference Datasets} & \multicolumn{2}{c|}{CMU1}                          & \multicolumn{2}{c|}{CMU2}                          & \multicolumn{2}{c|}{CMU3}                          & \multicolumn{2}{c|}{CMU4}                          & \multicolumn{2}{c}{Average}   \\ \hline
\textbf{Metrics}            & AP25          & \multicolumn{1}{c|}{mAP}           & AP25          & \multicolumn{1}{c|}{mAP}           & AP25          & \multicolumn{1}{c|}{mAP}           & AP25          & \multicolumn{1}{c|}{mAP}           & AP25          & mAP           \\ \hline
Dong et al.~\cite{dong2019fast}                        & 0.1           & \multicolumn{1}{c|}{39.9}          & 0.0           & \multicolumn{1}{c|}{29.2}          & 0.0           & \multicolumn{1}{c|}{29.2}          & 0.0           & \multicolumn{1}{c|}{32.8}          & 0.0           & 32.8          \\
VoxelPose~\cite{tu2020voxelpose}                   & 28.8          & \multicolumn{1}{c|}{86.9}          & 26.9          & \multicolumn{1}{c|}{69.6}          & 21.0          & \multicolumn{1}{c|}{69.6}          & 40.9          & \multicolumn{1}{c|}{87.9}          & 29.4          & 78.5          \\
MvP~\cite{zhang2021direct}                         & 0.0           & \multicolumn{1}{c|}{0.0}           & 0.0           & \multicolumn{1}{c|}{0.0}           & 0.0           & \multicolumn{1}{c|}{0.0}           & 0.0           & \multicolumn{1}{c|}{0.0}           & 0.0           & 0.0           \\
Ours                        & \textbf{86.8} & \multicolumn{1}{c|}{\textbf{96.1}} & \textbf{67.9} & \multicolumn{1}{c|}{\textbf{90.3}} & \textbf{52.5} & \multicolumn{1}{c|}{\textbf{78.8}} & \textbf{91.5} & \multicolumn{1}{c|}{\textbf{97.2}} & \textbf{74.7} & \textbf{90.6} \\ \hline
\end{tabular}

}
\setlength{\belowcaptionskip}{-12pt}
\caption{Generalization experiments on \textbf{changing camera arrangements}. All models are trained on CMU0, and tested on \textit{Inference Datasets}. The datasets CMU1-4 have different numbers and poses of cameras compared with CMU0.}
\label{table-generalization_arrangements}
\end{table*}

\begin{table*}[]
\centering

\resizebox{1.4\columnwidth}{!}{%
\begin{tabular}{@{}cc|cccccccc@{}}
\toprule
\multicolumn{2}{c|}{\textbf{Generalization}}                     & \multicolumn{8}{c}{Change scenes and datasets}                                                                                                     \\ \midrule
\multicolumn{2}{c|}{\textbf{Inference Datasets}}                 & \multicolumn{4}{c|}{Shelf}                                                         & \multicolumn{4}{c}{Campus}                                    \\ \midrule
\multicolumn{1}{c|}{Training Setting}              & Methods     & A1            & A2            & A3            & \multicolumn{1}{c|}{Average}       & A1            & A2            & A3            & Average       \\ \midrule
\multicolumn{1}{c|}{N/A}                           & Dong et al.~\cite{dong2019fast} & 98.8          & 94.1          & \textbf{97.8} & \multicolumn{1}{c|}{96.9}          & 97.6          & 93.3          & 98.0            & 96.3          \\ \midrule
\multicolumn{1}{c|}{\multirow{3}{*}{w/o Finetune}} & VoxelPose~\cite{tu2020voxelpose}   & 77.9          & 63.2          & 68.1          & \multicolumn{1}{c|}{69.8}          & 16.1          & 12.4          & 5.1           & 11.2          \\
\multicolumn{1}{c|}{}                              & MvP~\cite{zhang2021direct}         & 5.7           & 16.0          & 4.6           & \multicolumn{1}{c|}{8.7}           & 0.0           & 0.0           & 0.0           & 0.0           \\
\multicolumn{1}{c|}{}                              & Ours        & \textbf{89.9}          & \textbf{85.4}          & \textbf{88.6}          & \multicolumn{1}{c|}{\textbf{87.9}}          & \textbf{40.2}          & \textbf{61.0}          & \textbf{73.1}          & \textbf{58.1}          \\ \midrule
\multicolumn{1}{c|}{\multirow{3}{*}{w/ Finetune}}  & VoxelPose~\cite{tu2020voxelpose}   & 99.3          & 94.1          & 97.6          & \multicolumn{1}{c|}{97.0}          & 97.6          & 93.8          & \textbf{98.8} & \textbf{96.7} \\
\multicolumn{1}{c|}{}                              & MvP~\cite{zhang2021direct}         & 99.3          & 95.1          & \textbf{97.8} & \multicolumn{1}{c|}{97.4}          & 98.2          & 94.1          & 97.4          & 96.6          \\
\multicolumn{1}{c|}{}                              & Ours        & \textbf{99.5} & \textbf{96.8} & \textbf{97.8} & \multicolumn{1}{c|}{\textbf{98.0}} & \textbf{99.2} & \textbf{94.4} & 96.5          & \textbf{96.7} \\ \bottomrule
\end{tabular}
}

\setlength{\belowcaptionskip}{-12pt}
\caption{Generalization experiments on \textbf{changing scenes and datasets}. All learning-based models are trained on CMU0 from the CMU Panoptic dataset, and tested on \textit{Inference Datasets} with or without finetunning. Metric: PCP.}
\label{table-generalization_new_datasets}
\end{table*}

We present three types of generalization experiments with increasingly larger domain gaps including cross-cameras, cross-arrangements, and cross-datasets.
We follow Bartol et al.~\cite{bartol2022generalizable} to consider five camera arrangements with different camera numbers and camera poses on the CMU Panoptic dataset, and denote them as CMU0, CMU1, CMU2, CMU3, and CMU4, respectively. We show more details in Appendix~\ref{sec:supp:camera_arrangements}. 
We train all the models on the camera arrangement of CMU0 with 5 cameras. For in-domain experiments, we also test on CMU0. For generalization experiments, the models are tested on other arrangements and datasets without finetuning. 

\vspace{0.5em}
\noindent
\textbf{Cross-cameras} We remove or add a few cameras to CMU0. The results are in Table~\ref{table-generalization-camera_number}. ``CMU0($K$)'' means the test camera number is $K$. 
When removing cameras, all methods witness accuracy loss partly due to increased occlusion (see Figure \ref{fig:real_example_framework}for some examples).
However, the accuracies of VoxelPose and MvP drop more significantly than ours due to their weak generalization ability.

When we add new cameras, we observe that ours is the only method that can benefit from the additional information. For VoxelPose, the more cameras we add, the worse the result becomes. The accuracy of MvP even decreases to zero because the rays of the new cameras were not seen during training, making them hard to be correctly handled by the network. MvP tends to only ``memorize" the geometry for trained cameras but fails to generalize to unseen new cameras. 
Though the pure geometric method Dong et al.~\cite{dong2019fast} in theory has no generalization problem, it fails to accurately reconstruct poses on challenging CMU Panoptic dataset. Note that only qualitative results on CMU are given in the original paper. 
Ours achieves an obvious improvement over the baselines.

\vspace{0.5em}
\noindent
\textbf{Cross-arrangements} Then we evaluate in a more challenging setting by completely changing the camera arrangements, \eg having different camera numbers and poses from those used in training. The results are shown in Table~\ref{table-generalization_arrangements}.

Our method obtains notably better results than the baselines.
VoxelPose sacrifices accuracy a lot for new cameras.
MvP fails to work on unseen cameras.
The pure geometric method Dong et al.~\cite{dong2019fast}  fails to accurately reconstruct the poses due to severe occlusion.
Ours can be directly applied to CMU1 and CMU4 with high accuracy. 
The accuracy on the CMU2 and CMU3 datasets is relatively low due to the small number of cameras and the missing of the other side of the view, as shown in Appendix~\ref{sec:supp:camera_arrangements}. In summary, we achieve much higher accuracy on all arrangements compared with all the baselines.
Due to the difficulty of including all possible camera arrangements in training, the practical values of the other learning-based baselines are limited.

\vspace{0.5em}
\noindent
\textbf{Cross-datasets} 
We train the models on CMU0 and test them on the Shelf and Campus datasets. The results are in Table~\ref{table-generalization_new_datasets}.
When testing without finetuning, ours shows much better generalization performance than the baselines, where MvP still fails to generalize to new cameras. 
It is worth noting that ours achieves a PCP of 87.9\% on the Shelf.
Though much better than the learning baselines, the accuracy on Campus is relatively low due to the significant appearance domain gap and the larger outdoor space, which we will discuss more in Appendix~\ref{section:discuss}. 
Finetuning further improves the performance a lot. Ours achieves a state-of-the-art performance than both the learning and geometric baselines. 
Note that finetuning needs to collect training data in a new setting which is costly in time and resources, and sometimes not available. Ours with generalization capacity shows a potential trend to train once and deploy anywhere.
The pure geometric method Dong et al.~\cite{dong2019fast} gets reasonable results on the Shelf and Campus dataset. But as discussed in Table \ref{table-generalization-camera_number} and \ref{table-generalization_arrangements}, the performance drops when occlusion occurs.

\subsection{In-domain Performance}

We evaluate the in-domain performance of the models on three benchmarks, CMU Panoptic~\cite{Joo_2017_TPAMI}, Shelf~\cite{belagiannis20143d}, and Campus~\cite{belagiannis20143d} in Table~\ref{table-in_domain_experiments}. We use the same train and test splits following ~\cite{tu2020voxelpose,zhang2021direct}. 
The same camera arrangements are used for training and testing in this experiment. 
It demonstrates that the generalization ability of our method is achieved without sacrificing its fitting ability. Instead, it achieves comparable even better results than the state-of-the-art. We show qualitative results in Figure~\ref{fig:exp:cmu} and Appendix~\ref{sec:supp:visualizations}.

The results of the baselines are obtained from the papers (our re-implementation gets similar results). Our method outperforms VoxelPose~\cite{tu2020voxelpose} by a margin in terms of AP25, which suggests that the end-to-end framework can achieve precise estimation results by largely decreasing the quantization error. Besides, it performs comparably to or better than MvP~\cite{zhang2021direct}, which is also an end-to-end approach, on all metrics. However, as discussed in Section~\ref{section:generalization}, MvP has the risk of overfitting to the training cameras, making the results less meaningful. 
Since the Shelf and Campus datasets are small, we finetuned the model trained on CMU0 following the previous work~\cite{tu2020voxelpose,zhang2021direct}. Our method achieves comparable results to the state-of-the-art methods on both datasets. 
We have compared the generalization performance of ~\cite{tu2020voxelpose, zhang2021direct} and show a much better performance in Sec~\ref{section:generalization}.

\begin{table}[]
\centering
\resizebox{1.0\columnwidth}{!}{%
\begin{tabular}{l|cc|c|c}
\hline
\multicolumn{1}{c|}{\multirow{2}{*}{Methods}} & \multicolumn{2}{c|}{CMU Panoptic} & Shelf         & Campus        \\ \cline{2-5} 
\multicolumn{1}{c|}{}                         & AP25 $\uparrow$                & MPJPE $\downarrow$       & PCP $\uparrow$           & PCP $\uparrow$           \\ \hline
Belagiannis et al.~\cite{belagiannis20153d}                            & -                   & -           & 77.5          & 84.5          \\
Ershadi et al.~\cite{ershadi2018multiple}                                & -                   & -           & 88.0          & 90.6          \\
VoxelTrack~\cite{zhang2022voxeltrack}                                  & -                   & -           & 97.1          & 96.7          \\
Graph3D~\cite{wu2021graph}                                       & -                   & 15.8        & 97.7          & -             \\
Lin et al.~\cite{lin2021multi}                                       & 92.1                & 16.8        & 97.9          & 97.0          \\
Faster VoxelPose~\cite{ye2022faster}                                       & 85.2                & 18.3        & 97.6          & 96.2          \\
Chen et al.~\cite{chen2022vtp}                                       & 83.8                & 17.6        & 97.3          & 96.3          \\
Perez-Yus et al.~\cite{perez2022matching}                                       & -                   & -           & 96.5          & 96.7          \\
Dong et al.~\cite{dong2019fast}                                      & -                   & -           & 96.9          & 96.3          \\
VoxelPose~\cite{tu2020voxelpose}                           & 84.0                & 17.7        & 97.0          & 96.7          \\
MvP~\cite{zhang2021direct}                              & 92.3       & 15.8        & 97.4          & 96.6          \\
Ours                                          & \textbf{92.3}       & 16.0        & \textbf{98.0}          & 96.7          \\ \hline
\end{tabular}
}
\caption{In-domain experiments with the same training and testing settings. We consider the published image-based methods for a fair comparison. ``-" indicates missing data in the published paper.}
\label{table-in_domain_experiments}
    \vspace{-0.5cm}
\end{table}

\def\tablescaleabbl{0.88}
\begin{table*}
\centering
\begin{subtable}[c]{0.3\textwidth}
\centering

\resizebox{\tablescaleabbl\columnwidth}{!}{%
\begin{tabular}{@{}cccc@{}}
\toprule
Queries & \multicolumn{1}{l}{AP25} & \multicolumn{1}{l}{AP100} & \multicolumn{1}{l}{MPJPE} \\ \midrule
128    & 77.9                    & 93.4                     & 26.2                     \\
256    & 86.8                    & 97.4                     & 20.1                      \\
512    & 90.7                    & 98.4                     & 18.0                     \\
\textbf{1024}   & \textbf{92.3}           & \textbf{99.3}            & \textbf{16.0}            \\
1280   & 92.0                    & 99.6            & 16.0            \\ \bottomrule
\end{tabular}
}

\subcaption{Query number}
\label{table:abl:queries}
\end{subtable}
\begin{subtable}[c]{0.3\textwidth}
\centering
\resizebox{\tablescaleabbl\columnwidth}{!}{%
\begin{tabular}{@{}cccc@{}}
\toprule
Layers  & AP25           & AP100          & MPJPE          \\ \midrule
1 & 5.2 & 96.4 & 47.2 \\
2 & 79.2 & 97.9 & 19.9 \\
3 & 88.6 & 98.9 & 17.6 \\
\textbf{4} & \textbf{92.3} & \textbf{99.3} & \textbf{16.0} \\
5 & 91.0 & 99.4 & 16.3 \\
\bottomrule
\end{tabular}
}
\subcaption{Layer number}
\label{table:abl:layers}
\end{subtable}
\begin{subtable}[c]{0.3\textwidth}
\centering
\resizebox{\tablescaleabbl\columnwidth}{!}{%
\begin{tabular}{@{}cccc@{}}
\toprule
\multicolumn{1}{l}{KNN} & \multicolumn{1}{l}{AP25} & \multicolumn{1}{l}{AP100} & \multicolumn{1}{l}{MPJPE} \\ \midrule
1 & 86.0 & 98.5 & 19.1 \\
2 & 87.0 & 99.0 & 17.9 \\
3 & 90.5 & 99.2 & 17.4 \\
4 & 91.4 & 99.2 & 17.0 \\
\textbf{5} & \textbf{92.3} & \textbf{99.3} & \textbf{16.0} \\
6 & 90.8 & 99.1 & 16.7 \\ 
\bottomrule
\end{tabular}
}
\subcaption{Query assignment}
\label{table:abl:KNN}
\end{subtable}
\begin{subtable}[c]{0.3\textwidth}
\centering
\resizebox{\tablescaleabbl\columnwidth}{!}{%
\begin{tabular}{@{}cccc@{}}
\toprule
Feature      & AP25           & AP100          & MPJPE          \\ \midrule
Joint-level  & 90.9          & 99.2          & 17.0          \\
\textbf{Person-level} & \textbf{92.3} & \textbf{99.3} & \textbf{16.0} \\
Human-level  & 91.0              & 99.4              & 16.1              \\ \bottomrule
\end{tabular}
}
\subcaption{Appearance term}
\label{table:abl:query_design}
\end{subtable}
\begin{subtable}[c]{0.3\textwidth}
\centering
\resizebox{\tablescaleabbl\columnwidth}{!}{%
\begin{tabular}{@{}cccc@{}}
\toprule
Methods   & AP25           & AP100          & MPJPE          \\ \midrule
\textbf{MLP}       & \textbf{92.3} & \textbf{99.3} & \textbf{16.0} \\
Attention & 90.4          & 99.1          & 16.8         \\
Mean      & 36.2          & 54.0          & 71.8          \\ \bottomrule
\end{tabular}
}
\subcaption{Feature fusion}
\label{table:abl:feature_fusion}
\end{subtable}
\begin{subtable}[c]{0.3\textwidth}
\centering
\resizebox{\tablescaleabbl\columnwidth}{!}{%
\begin{tabular}{@{}cccc@{}}
\toprule
NMS & AP25           & AP100          & MPJPE          \\ \midrule
w/o & 20.0          & 20.8          & 16.2          \\
\textbf{w/}  & \textbf{92.3} & \textbf{99.3} & \textbf{16.0} \\ \bottomrule
\end{tabular}
}
\subcaption{NMS}
\label{table:abl:nms}
\end{subtable}

\setlength{\belowcaptionskip}{-12pt}
\caption{Ablation studies. The settings with the best AP25 have been highlighted in bold.}
\label{table-ablation}
\end{table*}

\subsection{Ablation Study}

\begin{figure*}
    \centering
    \includegraphics[width=0.9\textwidth]{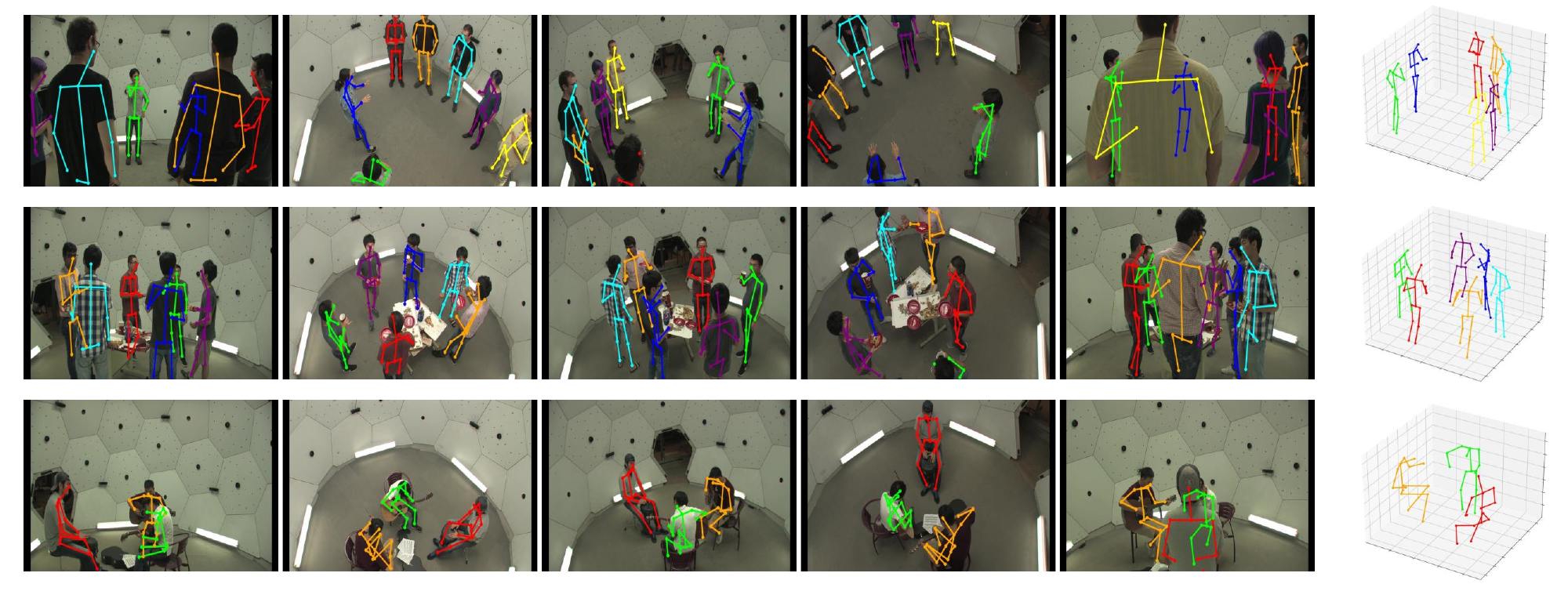}
    \vspace*{-4mm}
    \setlength{\belowcaptionskip}{-12pt}
    \caption{Sample estimation results on the CMU Panoptic dataset.}
    \label{fig:exp:cmu}
\end{figure*}

\noindent\textbf{Number of Queries} This parameter is mainly determined by the size of the motion capture space. The guideline is to make sure each GT pose has a sufficiently large overlap with at least one query so that the model can extract appropriate features for estimating the 2D poses in the first layer.  As shown in Table~\ref{table:abl:queries},
when the number of queries increases from 128 to 1024, the performance keeps growing and reaches the maximum performance at 1024. When we further increase the number of queries to 1280, the performance begins to slightly drop because it poses a challenge for the training. We didn't try the query numbers larger than 1280 because of the GPU memory limitation in training. 

\vspace{0.5em}
\noindent\textbf{Number of Decoder Layers} The 3D poses are iteratively refined through the decoder layers as shown in Figure \ref{fig:real_example_framework}. The improvement is significant in the first two layers and becomes marginal from the third layer. The observation agrees with the numerical results shown in Table~\ref{table:abl:layers}. When increasing the number of layers, the accuracy continuously improves and reaches the best with 4 layers, which validates that the proposed iterative refinement is effective. When we keep adding more layers, the performance saturates. It is worth noting that if we only use one layer, AP25 drops significantly to 6\%, showing that multiple decoder layers and iterative refinement are essential to obtain precise estimates. 

\vspace{0.5em}
\noindent\textbf{Query Assignment} During training, we select $W$ queries whose initialized geometry terms are nearest to the ground truth poses. Then, we enforce pose losses to train those queries to output the matched groundtruth poses. As shown in Table~\ref{table:abl:KNN}, when setting $W$ to 5, the performance of AP25 improves from $86.0\%$ to $92.3\%$. Compared with the one-to-one matching strategy, e.g., Hungarian loss in DETR~\cite{carion2020end}, the multiple-to-one matching strategy can augment the training data and help the network converge better. The performance begins to drop when selecting more than 5 queries since some queries far away are also matched which is difficult to learn. In particular, we tried the Hungarian loss but it did not converge in our experiments.

\vspace{0.5em}
\noindent\textbf{Appearance Term} We evaluate three strategies for appearance term embedding. The first is the joint-level strategy where each joint in each query has independent parameters. The second is human-level embedding where all queries share the same parameters. The third is our adopted hierarchical embedding strategy. The results are shown in Table~\ref{table:abl:query_design}. The human-level and person-level strategies achieve better results than the joint-level.  Compared with a human-level feature, person-level is more flexible and each query can learn a feature related to its initialized positions and achieves the best performance. 

\vspace{0.5em}
\noindent\textbf{Feature Fusion} In each decoder layer, we need to update the queries by fusing the features from multiple views. We ablated several fusion methods in Table~\ref{table:abl:feature_fusion}. We notice that MLP achieves the best results, even a little better than attention. We think that attention has better capacity but also is harder to learn compared with MLP. 

\vspace{0.2em}
\noindent\textbf{Non-maximum Suppression} As we use the KNN-based instead of the one-to-one matching strategy during training, multiple queries may have large confidence scores for a single GT pose. Non-maximum suppression is important to remove the redundant estimates, as shown in Figure~\ref{fig:distributions_queries}. Without NMS, we can see that AP25 and AP100 drop significantly in Table~\ref{table:abl:nms}. When using NMS to filter out those redundant estimates, the accuracy is boosted to $92.3\%$. 

\vspace{0.2em}
\noindent\textbf{Inference Speed}
We use a single V100 GPU for inference. Ours needs 0.21s for each inference, which is faster than VoxelPose, which needs 0.29s. 
Ours lies inside a 19\% range with MvP which needs 0.17s. A more efficient query sampling strategy remains in future work.

\vspace{0.2em}
We provide more ablation studies in the Appendix.

\section{Conclusion}

We propose a hybrid method for multi-view 3D human pose estimation, combining a classical geometry module with a learned appearance module in a coarse-to-fine iterative network. Differing from the previous methods, we explicitly decouple 2D appearance and 3D geometry in the Transformer architecture to achieve strong generalization performance and yet maintain the robustness benefits of end-to-end learning. The 2D module directly estimates and refines the 2D poses from the images, while the 3D module recovers the 3D poses using triangulation and is immune to quantization errors. This approach results in an end-to-end trained model that generalizes well across different cameras and datasets. 
Our framework is generalizable and can be applied to other keypoint estimation tasks such as shape, hand, and face estimation.

{
    \small
    \bibliographystyle{ieeenat_fullname}
    \bibliography{main}
}

\clearpage
\setcounter{page}{1}
\maketitlesupplementary

\newcommand{\beginsupplement}{
  \setcounter{table}{0}  
  \renewcommand{\thetable}{A\arabic{table}} 
  \setcounter{figure}{0} 
  \renewcommand{\thefigure}{A\arabic{figure}}
  \setcounter{section}{0}
  \renewcommand{\thesection}{A\arabic{section}}
}

\beginsupplement

We present more implementation details and experiment results in this supplementary.

\section{Network Architectures}
\label{sec:suppl:network_architectures}

There are several MLP networks used in our model. We present the detailed structures here. The feature dimension of the appearance term is $256$. 

\paragraph{Network 1: $g_\theta (.)$} This network is used for estimating the 2D pose residuals in the appearance module. It has three layers with their hidden dimensions being set to $256$. We use ReLU as the non-linear activation layer.

\paragraph{Network 2: $f_\alpha(.)$} This network is used for feature fusion as shown in Equation 4. It is an MLP with one linear layer that maps the appearance term to a vector with the same dimension.

\paragraph{Network 3: $f_\gamma(.)$} This network is also used for feature fusion as shown in Equation 4. It has two layers with their hidden dimensions being set to 1024.  We use ReLU as the non-linear activation layer. There is also a residual branch. We also use layer norm.

\paragraph{Network 4: $f_\beta(.)$} This network is for query classifier. It is a linear layer that maps a vector with a dimension of $L$  to two scores followed by a sigmoid function with the output representing the positive and negative probability, respectively.  

\section{Visualizations}
\label{sec:supp:visualizations}

\begin{figure*}[h]
    \centering
    \includegraphics[width=1.0\textwidth]{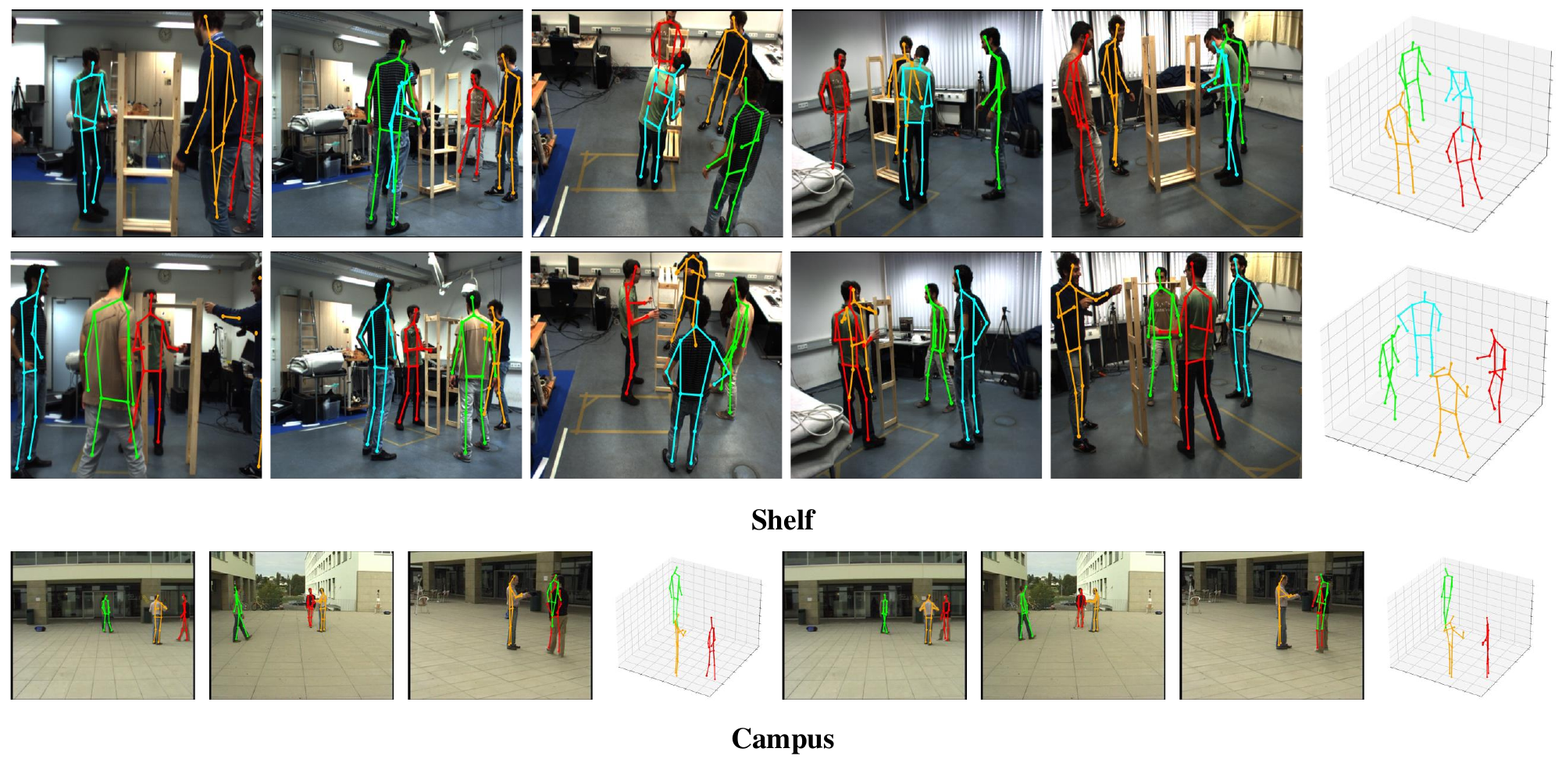}
    \caption{Sample results on Shelf and Campus datasets.}
    \label{fig:exp:shelf_campus}
\end{figure*}

\textbf{Qualitative results on Shelf and Campus}. We have visualized some of the estimated results on the images from the CMU Panoptic dataset in Figure~\ref{fig:exp:cmu}. We also visualize the results on the images from the Shelf and Campus datasets, in Figure~\ref{fig:exp:shelf_campus}. Our method can accurately estimate the poses for people in different postures even when severe occlusion occurs in some views. While the results are nearly perfect on the Shelf dataset, we notice that in the Campus dataset, our method gets inaccurate estimates for the person in a pink shirt (lower arm, yellow skeleton). This usually occurs when the number of cameras is small and the body joints are occluded in most views. We discuss the limitation in Sec~\ref{section:discuss}.

\begin{figure}
    \centering
    \includegraphics[width=0.48\textwidth]{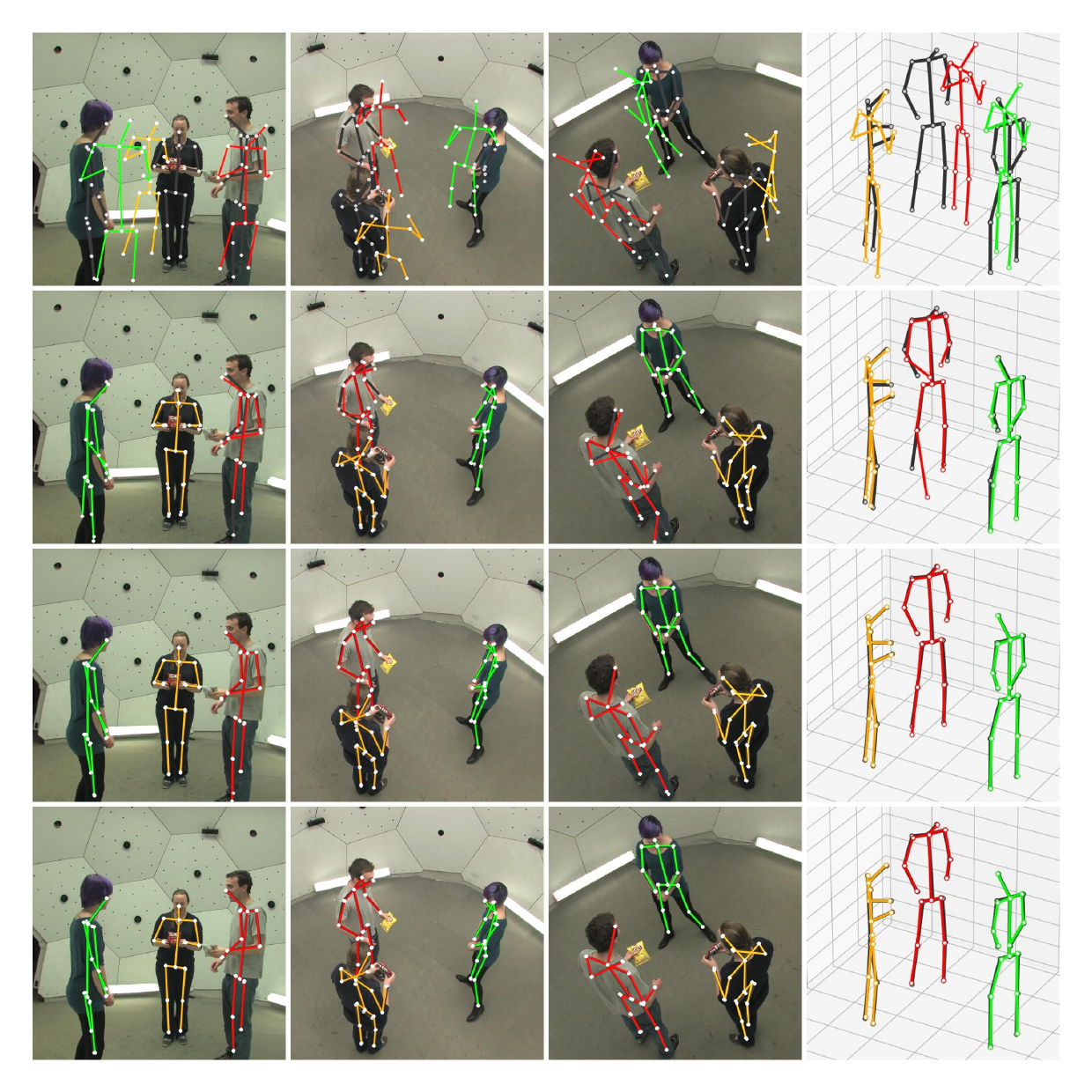}
    \caption{Progressive refinement of the 2D and 3D poses over four decoder layers. The outputs from Layer 1 to Layer 4 are shown from top to bottom. The images of Camera 1 to Camera 4 and the 3D pose are shown from left to right.
    Black and colorful skeletons denote the GT and estimated poses, respectively. }
    \label{fig:supp:decoders}
\end{figure}

\paragraph{Results from Each Decoder Layer} We visualize the 2D and 3D pose estimations of 3 persons on the CMU Panoptic dataset from each decoder layer in Figure~\ref{fig:supp:decoders}. The geometry queries are refined from coarse initializations to accurate poses through four transformer decoder layers. 

\section{Camera Arrangements}
\label{sec:supp:camera_arrangements}
\begin{figure*}
    \centering
    \includegraphics[width=0.98\textwidth]{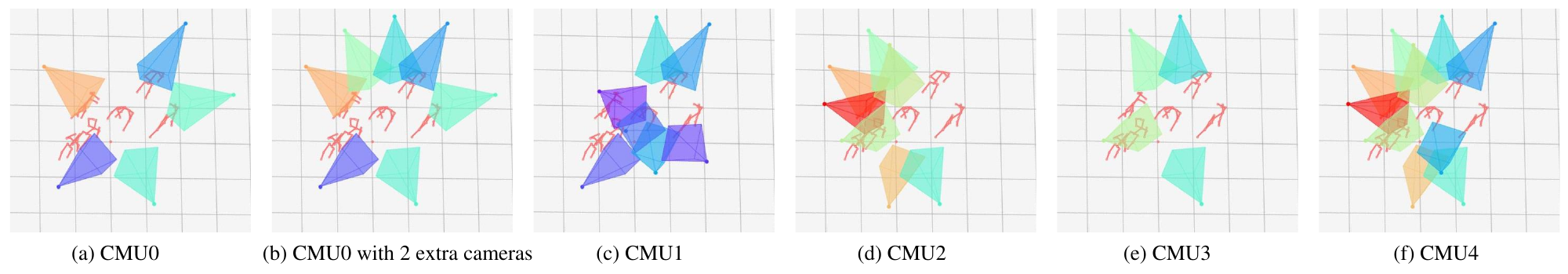}
    \caption{Visualization of the camera arrangements of sequences CMU0 to CMU4.}
    \label{fig:cmu_settings}
\end{figure*}

We use different camera arrangements following \cite{bartol2022generalizable}. We show the details of the camera arrangements of CMU0-CMU4 used in the generalization experiments in Table~\ref{table:camera_arrangements}, and visualize them in Figure~\ref{fig:cmu_settings}. 
Among them, CMU2 and CMU3 are more challenging because they either have a small number of cameras, or the cameras miss the other side of views.

\begin{table*}[]
\centering
\begin{tabular}{@{}llc@{}}
\toprule
Camera Arrangements & Camera IDs                           & \multicolumn{1}{l}{Camera Num} \\ \midrule
CMU1                & 1, 2, 3, 4, 6, 7, 10                 & 7                              \\
CMU2                & 12, 16, 18, 19, 22, 23, 30           & 7                              \\
CMU3                & 10, 12, 16, 18                       & 4                              \\
CMU4                & 6, 7, 10, 12, 16, 18, 19, 22, 23, 30 & 10                             \\ \midrule
CMU0                & 3, 6, 12, 13, 23                     & 5                              \\
CMU0 w/ 2 extra cameras             & 3, 6, 12, 13, 23, 10, 16             & 7                              \\
CMU0(K)             & First K cameras in CMU0 w/ 2           & K                              \\ \bottomrule
\end{tabular}
\caption{The details of the camera arrangements.}
\label{table:camera_arrangements}
\end{table*}

\section{More Ablation Studies}

\paragraph{Number of Cameras} To further investigate the influence of the number of cameras, we train the model on CMU0(7) with two extra cameras on CMU0 as can be seen in Figure \ref{fig:cmu_settings} (b). Then we gradually decrease the number of cameras and evaluate their performance in Table~\ref{table:abl:camera_num}. 
Our method is consistently better than VoxelPose, which validates its strong generalization performance. More importantly, we can see that AP100 barely changes when we decrease the number of cameras from 7 to 4. In extreme cases, when we only have two cameras available, AP100 of our method is still reasonable considering the severe occlusions in the dataset.  
It may be helpful to note that this experiment is different from Table 1 in the main paper where the model is trained on five cameras CMU0(5).

\begin{table*}[hbt!]
\centering
\begin{tabular}{@{}c|ccc|ccc@{}}
\toprule
\multirow{2}{*}{Cam Num} & \multicolumn{3}{c|}{VoxelPose~\cite{tu2020voxelpose}} & \multicolumn{3}{c}{Ours} \\ \cmidrule(l){2-7} 
                         & AP25     & AP100    & MPJPE    & AP25            & AP100   & MPJPE  \\ \midrule
7                        & 89.8     & 98.3     & 16.1     & \textbf{95.3}   & 99.5    & 14.6   \\
6                        & 87.1     & 98.0     & 17.1     & \textbf{94.5}   & 99.4    & 15.2   \\
5                        & 68.7     & 84.8     & 18.9     & \textbf{91.1}   & 99.2    & 17.0   \\
4                        & 35.8     & 80.8     & 25.1     & \textbf{75.6}   & 98.9    & 21.0   \\
3                        & 1.8      & 40.6     & 67.4     & \textbf{35.1}   & 94.3    & 36.3   \\
2                        & 0.0      & 2.1      & 164.7    & \textbf{1.9}    & 35.7    & 97.0   \\ \bottomrule
\end{tabular}
\caption{Ablation study on the number of cameras. Both models are trained on CMU0(7).}
\label{table:abl:camera_num}
\end{table*}

\paragraph{Loss weights of decoder layers} As shown in Table~\ref{table:abl:weightdecay}, ``All layers" means adding loss to all four layers. ``w/o decay" means using the same loss weight for each layer without weight decay. ``Exp decay" means the loss weights decaying exponentially as $1,0.5,0.25,0.125$ for the last layer to the first layer, respectively. The idea is to tolerate the errors in the earlier layers. ``Linear decay" means the loss weights decaying linearly as $1,0.75,0.5,0.25$ for the last layer to the first layer, respectively. We get the best performance when adding the same weight loss to all the layers. The performance decreases dramatically when only adding loss to the final layer, denoted by ``Final layer", which shows it is important to add a supervision signal for each layer for good convergence.

\begin{table}[t!]
\centering
\begin{tabular}{@{}lccc@{}}
\toprule
Configurations             & \multicolumn{1}{l}{AP25} & \multicolumn{1}{l}{AP100} & \multicolumn{1}{l}{MPJPE} \\ \midrule
All layers w/o decay       & \textbf{92.3}           & \textbf{99.3}            & \textbf{16.0}            \\
All layers w/ Exp decay    & 91.2                    & 99.0                        & 17.2                     \\
All layers w/ Linear decay & 90.4                    & 99.0                     & 17.0                        \\
Only final layer                 & 0.2                     & 57.0                     & 96.1                     \\ \bottomrule
\end{tabular}
\caption{Adding weight decay among decoder layers.}
\label{table:abl:weightdecay}
\end{table}

\paragraph{Denoise Training} In the main experiments, we use uniform sampling to initialize geometry queries during both training and inference. Here we explore another method for initialization during training. In particular, we add noises to the ground truth poses with a sigma of $\sigma$ and use the noised poses as the initialized queries. This method is denoted by ``GT Noise $\sigma$''. During inference, we use the output of VoxelPose to initialize the queries. As shown in Table~\ref{table:abl:denoisetraining}, our system can boost up the AP25 of VoxelPose from 85.3\% to 92.7\% under ``GT Noise 20'', which is also slightly higher than the sampling-based initialization. It proves that our system has a higher accuracy when given an accurate initialization, and can act as a refiner to further improve the performance of other human pose estimation methods.

\begin{table}[t!]
\centering
\begin{tabular}{@{}lllc@{}}
\toprule
Model      & Training Init & Infer Init & \multicolumn{1}{l}{AP25} \\ \midrule
VoxelPose  & /             & /          & 85.3                     \\
Ours & GT Noise 200  & VoxelPose  & 90.3                     \\
Ours & GT Noise 100  & VoxelPose  & 91.4                     \\
Ours & GT Noise 50   & VoxelPose  & 92.4                     \\
Ours & GT Noise 20   & VoxelPose  & \textbf{92.7}            \\
Ours & Sampling        & Sampling     & 92.3                     \\ \bottomrule
\end{tabular}
\caption{Denoise training experiments. During training, we initialize with noisy ground truth poses; during inference, we initialize with VoxelPose. We can further improve the results of VoxelPose.}
\label{table:abl:denoisetraining}
\end{table}

\begin{table}[]
\centering
\begin{tabular*}{0.48\textwidth}{@{\extracolsep{\fill}}lccc}
\toprule
Decoder Layer & \multicolumn{1}{l}{AP25} & \multicolumn{1}{l}{AP100} & \multicolumn{1}{l}{MPJPE} \\ \midrule
Independent   & \textbf{92.3}           & \textbf{99.3}            & \textbf{16.0}            \\
Sharing       & 85.5                    & 98.1                     & 19.0                     \\ \bottomrule
\end{tabular*}
\caption{Ablation on parameter sharing for decoder layers.}
\label{table:abl:sharingweights}
\end{table}

\noindent\textbf{Parameters Sharing} Since all decoder layers share the same goal of refining the 2D and 3D poses, we evaluate whether we can use the same parameters for them which can reduce the number of parameters. As shown in Table~\ref{table:abl:sharingweights}, if we share the parameters, the performance decreases significantly in terms of AP25. But for AP100, it barely changes. The results suggest that using independent parameters is helpful to improve estimation precision. The front and last layers tend to learn different focuses for coarse and fine updates.

\paragraph{The MLP in Feature Fusion} As shown in Table~\ref{table:abl:fgama}, when adding the MLP $f_\gamma$ in the \textit{feature fusion} process, the accuracy can be further improved.

\begin{table}[t!]
\centering
\begin{tabular}{@{}lccc@{}}
\toprule
FFN & \multicolumn{1}{l}{AP25} & \multicolumn{1}{l}{AP100} & \multicolumn{1}{l}{MPJPE} \\ \midrule
w/  & \textbf{92.3}           & \textbf{99.3}            & \textbf{16.0}            \\
w/o & 88.6                    & 99.0                       & 17.6                     \\ \bottomrule
\end{tabular}
\caption{Ablation study of the MLP $f_\gamma(.)$ in Feature Fusion.}
\label{table:abl:fgama}
\end{table}

\section{Camera Parameters}
\label{sec:suppl:camera_parameters}

Camera parameters including the extrinsic poses and intrinsic matrix are needed for triangulation. Camera parameters are practical to get in real applications. 
For scenarios like surveillance where cameras are fixed, a one-time camera calibration~\cite{salvi2002comparative} can provide the poses and intrinsic. 
There are also online calibration methods~\cite{puwein2015joint,lee2022extrinsic} to estimate camera poses dynamically. 
Since the camera calibration is well studied in the literature and is out of the scope of this paper, we assume the camera pose is known and use the provided parameters in the datasets following the common practices in the literature~\cite{zhang2021direct, tu2020voxelpose}. All the baselines assume the camera poses are provided, so we have a fair comparison to them and can highlight the performance improvement brought by our model generalization ability. 

\section{Computation Analysis}

We show parameter counts, flops (evaluated by MACs), and running times in Table~\ref{tab:computation}. Ours has the fewest parameters among the 3 methods and comparable MACs with MvP. We use a single V100 GPU for inference. Ours needs 0.21s for each inference, which is faster than VoxelPose, which needs 0.29s. Ours lies inside a 19\% range with MvP which needs 0.17s. A more efficient query sampling strategy remains in future work.

\begin{table}[]
    \centering
    \begin{tabular}{@{}lccc@{}}
    \toprule
              & \multicolumn{1}{l}{Param (M)} & \multicolumn{1}{l}{MACs (G)} & \multicolumn{1}{l}{Time (ms)} \\ \midrule
    MvP       & 42.6                          & \textbf{567.0}                 & \textbf{169.5}                \\
    VoxelPose & {\ul 40.6}                    & 972.2                        & 291.3                         \\
    Ours      & \textbf{37.6}                 & {\ul 639.7}                  & {\ul 205.1}            \\ \bottomrule
    \end{tabular}
    \caption{Computation analysis.}
    \label{tab:computation}
\end{table}

\section{Limitation and Future Work}
\label{section:discuss}

As can be seen from the visualizations in Figure \ref{fig:exp:shelf_campus} (the lower arm of the yellow skeleton on Campus dataset is inaccurate), our method suffers when the number of camera views is extremely small and the body joints are severely occluded in most views. This is because our system relies on triangulation which requires accurate 2D positions in at least two views to recover the 3D position accurately. Note that this also poses challenges for other methods and our method actually performs better than them. Some methods such as VoxelPose are slightly more robust to occlusion because they use 3D convolutions to mix the features of all joints, which allows to make coarse predictions for the occluded joints based on the visible ones. Inspired by that, one possible way to enhance our method is to use robust structural triangulation ~\cite{chen2022structural} instead of the current keypoint-wise triangulation ~\cite{iskakov2019learnable} to explore the dependency among all joints to help estimate the occluded joints. Besides, we can take advantage of the constraints from the scenes and human interactions~\cite{hassan2021populating} to further improve the accuracy. Finally, it will be interesting to extend the Transformer architecture into a video-based system~\cite{zhang2022voxeltrack,reddy2021tessetrack,zhang20204d,dong2021fast} which further fuses temporal information for robust tracking.


\end{document}